\documentclass{article}
\usepackage[letterpaper,top=2cm,bottom=2cm,left=3cm,right=3cm,marginparwidth=1.75cm]{geometry}

\usepackage{graphicx}

\usepackage{tikz}
\usepackage{comment}
\usepackage{amsmath,amssymb,amsthm} 
\usepackage{color}

\usepackage{xspace}
\usepackage{amssymb}
\usepackage{booktabs}
\usepackage{comment}
\usepackage{xcolor}
\usepackage{multirow}
\usepackage{multicol}
\usepackage{enumerate}
\usepackage{nccmath}

\DeclareMathOperator*{\argmax}{argmax}
\DeclareMathOperator*{\argmin}{argmin}

\newcommand{\alg}{\textit{Essential Features}\xspace}
\newcommand{\etal}{et al.\xspace}
\newtheorem{theorem}{Theorem}

\title{\textbf{Content-Adaptive Pixel Discretization to Improve Model Robustness}}

\author{\Large{Ryan Feng${}^1$, Wu-chi Feng${}^2$, Atul Prakash${}^1$}\vspace{0.1em}\\
\normalsize{${}^1$University of Michigan, ${}^2$Portland State University}\vspace{0.1em}\\
\normalsize{${}^1$\{rtfeng, aprakash\}@umich.edu, ${}^2$\{wuchi\}@pdx.edu
}}

\date{}

\begin{document}

\maketitle

\begin{abstract}
Preprocessing defenses such as pixel discretization are appealing to remove adversarial attacks due to their simplicity. However, they have been shown to be ineffective except on simple datasets like MNIST. We hypothesize that existing discretization approaches failed because using a fixed codebook for the entire dataset limits their ability to balance image representation and codeword separability. We first formally prove that adaptive codebooks can provide stronger robustness guarantees than fixed codebooks as a preprocessing defense on some datasets. Based on that insight, we propose a content-adaptive pixel discretization defense called \alg, which discretizes the image to a per-image adaptive codebook to reduce the color space. We then find that \alg can be further optimized by applying adaptive blurring before the discretization to push perturbed pixel values back to their original value before determining the codebook. Against adaptive attacks, we show that content-adaptive pixel discretization extends the range of datasets that benefit in terms of both $L_2$ and $L_\infty$ robustness where previously fixed codebooks were found to have failed. Our findings suggest that content-adaptive pixel discretization should be part of the repertoire for making models robust. 
\end{abstract}

\begin{figure}[t]
     \centering
       {\makebox[4.5in]{\includegraphics[width=1in]{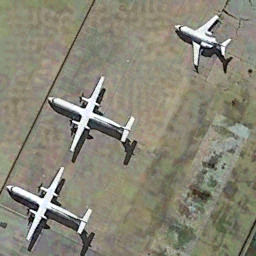}\hfill\includegraphics[width=1in]{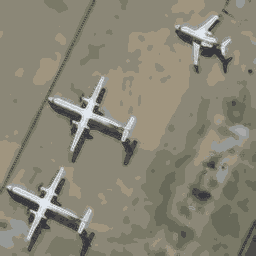}\hfill\includegraphics[width=1in]{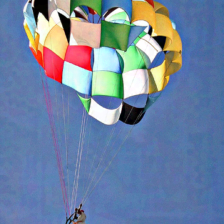}\hfill\includegraphics[width=1in]{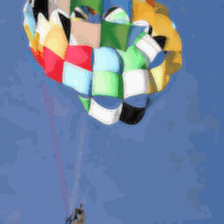}}}%
    \hfill
    \caption{\alg applies adaptive blurring and content-adaptive pixel discretization to constrain the attack space to a smaller color space. Left of each pair: attacked image. Right of each pair: transformed versions of the attacked image.}
    \label{fig:intro}
    \vspace{-0.1in}
\end{figure}

\section{Introduction}
Machine learning models have been used for a large diversity of tasks including robots, 
speech recognition systems, and 
self-driving cars. These models, however, have been shown to be vulnerable to subtle adversarial attacks~\cite{athalye2017synthesizing,carlini2017towards,roadsigns17,goodfellow2014explaining,kurakin2016adversarial,szegedy2016rethinking}, 
threatening the safety and real-world practicality of such systems.


Image preprocessing techniques such as JPEG compression~\cite{das2017keeping,dziugaite2016study,guo2017countering}, color-bit reduction~\cite{guo2017countering,xu2017feature}, and blurring~\cite{xu2017feature} were proposed as early approaches towards defending image classification models against adversarial attacks. Such techniques were appealing because they 
were computationally cheap and simple, with the intuition being 
they could
reduce adversarial effects without drastically changing the appearance of the image. These defenses, however, were later defeated with adaptive attacks that accounted for the transform~\cite{athalye2018obfuscated}.

Chen et al.~\cite{chen2019towards} argue that pixel discretization is unlikely to ever provide robustness on all but the simplest datasets. To back this claim, they extract theoretical and empirical insights on a dataset specific fixed codebook approach, where a codebook is a set of allowed codeword colors. They found that their technique succeeded on simple datasets like MNIST but failed on more practical datasets due to a lack of globally-separated color clusters. 

We observe that such discretization techniques must inherently trade-off \emph{representation}, the ability to preserve the essential features of the natural image required for classification, and \emph{separability}, the scale of the space between codeword colors. Particularly with more complex datasets such as CIFAR-10~\cite{krizhevsky2009learning} that contain more colors in the natural data, codebooks with fewer colors tend to not have enough colors to represent the natural images well enough while codebooks with more colors will have less separation between the codewords.

Our key insight is that we can improve the representation-separability trade-off with an \emph{adaptive} transformation tailored to the content of each image. Individual images may have good color separability even if the overall dataset does not (i.e., there does not exist a single codebook that suffices for all images). For instance, a STOP traffic sign is mostly red and white and has good separability. Similarly, an image of a flamingo against a water background may have good separability, but may require the inclusion of pink, which could be close to red if using a single codebook for both. 

We first formally prove that adaptive codebooks can provide stronger robustness guarantees than fixed codebooks on some datasets (Section~\ref{sec:motivation}). To do so, we prove that adaptive codebooks can provide a guaranteed robustness on a generalized version of MNIST that includes digits at multiple different color intervals rather than primarily just white and black. This proof shows a concrete example where fixed codebooks can struggle if different inputs require too many different sets of colors.

Based on this insight, we propose a content-adaptive pixel discretization defense (or \emph{adaptive discretization} for short) using per-image adaptive codebooks (Section~\ref{sec:approach}). To find such codebooks, we apply a variant of $k$-means color reduction, where $k$ is chosen adaptively and tailored to each image. 

However, the adversary does have limited influence over an image's adaptive codebook, since that codebook is based on perturbed image content. We find that adaptive discretization can be often improved by first applying an edge-aware adaptive blurring to bias perturbed pixels back towards their true clusters without destroying important edge features. We term the combined approach \alg, as it attempts to preserve key edges and colors.

On complex datasets with fewer dataset-wide color trends, we empirically find that applying \alg with adversarial training generally increases the $L_2$ robustness compared against adversarial training baselines (e.g., 20.15\% to 38.46\% for CIFAR-10~\cite{krizhevsky2009learning} and 12.09\% to 52.44\% for RESISC45~\cite{cheng2017remote}) (Section~\ref{sec:experiments}). We also find our approach to be comparable to fixed codebook approaches on datasets with clearer codebook choices. \alg also results in substantial increases in the $L_\infty$ robustness on several datasets (e.g., 43.89\% to 53.33\% on RESISC45 and 75.97\% to 80.71\% on Imagenette~\cite{imagenette}), suggesting that adaptive discretization can be a useful technique for model robustness where fixed codebooks fail. Finally, we discuss the results and their limitations (Section~\ref{sec:analysis}).

\section{Related Work}
\textbf{Image Preprocessing Defenses:} 
Several preprocessing defenses include color-bit reduction~\cite{guo2017countering,xu2017feature}, JPEG compression~\cite{das2017keeping,dziugaite2016study,guo2017countering}, and a non-differentiable pixel deflection approach~\cite{prakash2018deflecting}.
These image preprocessing algorithm defenses were broken with simple applications of BPDA with the identity function as the backwards approximation~\cite{athalye2018robustness,athalye2018obfuscated}. Xu \etal additionally propose filtering defenses but do not present adaptive attacks. Liang \etal propose a per image adaptive detection approach that blurs each image based on its entropy~\cite{liang2018detecting}. In contrast, we propose a defense that outputs correct predictions in the face of attacks and blurs adaptively within an image.

Chen \etal propose a dataset aware color codebook discretization algorithm that reduces each image to the same set of separable codebook colors~\cite{chen2019towards} but suggest that such techniques are doomed on complex datasets such as ImageNet due to a lack of color separation in the dataset. We address this by instead extracting codebooks on a \emph{per-image basis}.

Jalalpour \etal~\cite{jalalpour2019leveraging} proposed the notion of using $k$-means color reduction to thwart adversarial attacks. We also use a variant of $k$-means color reduction; however, unlike Jalalpour \etal, we choose the number of colors $k$ adaptively per image such that the chosen clusters are highly separable. Jalalpour \etal also did not adversarially train networks with $k$-means color reduced images. Finally, we add adaptive Gaussian blurring before color reduction rather than normal Gaussian blurring, which better preserves the original edge features.

\noindent\textbf{Adversarial Training:} Adversarial training is a popular 
defense that trains classifiers on attacked images to raise robustness. A common approach is Madry Adversarial training~\cite{madry2017towards}, which trains classifiers on PGD~\cite{madry2017towards} attacks. While effective, one limitation of adversarial training is that it is slow and hard to scale~\cite{kurakin2016scale}, so other approaches have tried to speed up adversarial training while retaining comparable accuracy~\cite{shafahi2019adversarial,wong2020fast,zhang2019you,zhang2019theoretically,zheng2020efficient}. We use ATTA~\cite{zheng2020efficient}, a speed-up technique that leverages cross-epoch transferability. 


\section{Motivation}\label{sec:motivation}



 
 
 To motivate our approach, we show a scenario where an adaptive codebook strategy offers provably stronger robustness guarantees than a fixed codebook strategy that uses the same codebook for every input image, implying that adaptive codebooks can outperform fixed codebooks for some datasets.

 \subsection{Preliminaries}
\textbf{Fixed Codebook Function ($FC_C$):} Let $FC_C : \mathbb{Z}^{H \times W} \to \mathbb{Z}^{H \times W}$ be a function that takes an input image $x$ consisting of a single-channel (grayscale) integer colors in $[0..255]$ and discretizes each pixel $x[i, j]$ for $0 \leq i < H$ and $0 \leq j < W$ to the value of the nearest color in the given codebook of $n$ ordered colors $C = \{c_1, c_2, ..., c_n\}$, where each color is a distinct integer in $[0..255]$. Ties are broken by assigning the pixel $x[i, j]$ to the smaller color, but the proof can easily be adapted if they go to the larger color. 
 

\noindent\textbf{MNIST$_G$ Dataset:} Let our dataset be a generalized form of the MNIST test set called MNIST$_G$, consisting of discretized images set to all possible pairs of colors $w$ or more apart for some $w \geq 5$ (see Supplement Section~\ref{sec:supp_proof}). Denote $(x, y)$ as an image $x$ with label $y$. We derive MNIST$_G$ by first discretizing the original MNIST test set using a fixed codebook with two colors of black (0) and white (255), resulting in a dataset called MNIST$_D$, and then remapping the colors of 0 and 255 in  MNIST$_D$ to all pairs of integer colors $c_1$ and $c_2$ that are at least $w$ apart. The intermediate set $A$ denotes the possible tuples of parameters:

\begin{equation}
    \text{MNIST}_D = \{(FC_{\{0, 255\}}(x), y) : (x, y) \in \text{MNIST}\}
\end{equation}

\begin{equation}
    A = \{((x, y), c_1, c_2) \in \text{MNIST}_D \times [0..255] \times [0..255] : c_2 - c_1 \geq w\}
\end{equation}

\begin{equation}
    \text{MNIST}_G = \{(FC_{\{c_1, c_2\}}(x), y) : ((x, y), c_1, c_2) \in A\}
\end{equation}

\noindent\textbf{Model:} Let $M : \mathbb{Z}^{H \times W} \to \mathbb{Z}$ be a model from prior work~\cite{chen2019towards} that achieves high natural accuracy on MNIST images discretized to \{0, 255\}, i.e., on MNIST$_D$.

\noindent\textbf{Adaptive Codebook Function (AC):} Let $AC : \mathbb{Z}^{H \times W} \to \mathbb{Z}^{H \times W}$ refer to a function that extracts an adaptive codebook of two colors, consisting of the following algorithm for a given input image $x$:

\begin{enumerate}
    \item Find $(i^*_1, j^*_1), (i^*_2, j^*_2)$ = $\argmax_{(i_1, j_1) \in x, (i_2, j_2) \in x} x[i_2, j_2] - x[i_1, j_1]$, or in other words, the pair of pixels with the largest difference in $x$.
    \item Set $c_{mid} = \frac{x[i^*_1, j^*_1] + x[i^*_2, j^*_2]}{2}$ (the precise float value, not integer truncated).
    \item Assign all pixels in the input image $x$ such that $x[i, j] \leq c_{mid}$ to $x[i^*_1, j^*_1]$ and all pixels in the input image $x$ such that $x[i, j] > c_{mid}$ to $x[i^*_2, j^*_2]$.
\end{enumerate}






\noindent\textbf{Cluster Assignment Property:} We now define the Cluster Assignment Property (CAP) below. Given an input tuple of an image $x$, perturbation $\delta$, and transform function $T$, the tuple satisfies the property if for all pairs of pixels, the pair matching each other in color in $x$ implies that their color will also match in $T(x + \delta)$. In other words, groupings of equally valued pixels in $x$ are preserved:

\begin{equation}\label{eq:CAP}
    \forall i_1 \forall j_1 \forall i_2 \forall j_2 \hspace{0.5em} x[i_1, j_1] = x[i_2, j_2] \implies T(x + \delta)[i_1, j_1] = T(x + \delta)[i_2, j_2]
\end{equation}

\noindent\textbf{Remapping Function (S):} Let $S: \mathbb{Z}^{H \times W}\times \mathbb{Z}^{H \times W}$ be a function that takes an image consisting of exactly two colors and maps the lower color to 0 and the higher color to 255.

\begin{figure}[t]
     \centering
       {\makebox[4.5in]{\includegraphics[width=0.7in]{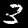}\hfill\includegraphics[width=0.7in]{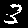}\hfill\includegraphics[width=0.7in]{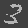}\hfill\includegraphics[width=0.7in]{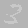}}}%
    \hfill
    \caption{MNIST$_G$ example. Left to right: 1) MNIST image. 2) MNIST$_D$ version. 3) One such MNIST$_G$ image where a codebook of \{0, 255\} would still recover the MNIST$_D$ image. 4) One such MNIST$_G$ image where a codebook of \{0, 255\} would set everything to 255.}
    \label{fig:mnistg}
    \vspace{-0.1in}
\end{figure}

Fig.~\ref{fig:mnistg} shows an intuitive example of MNIST$_G$ explaining why a fixed codebook (e.g.,  of two colors 0 and 255) will fall short in some cases.

\subsection{Theorems and Proof Sketches}
\begin{theorem}\label{thm:poc1}
Given any arbitrary image $x \in $ MNIST$_G$, any arbitrary integer perturbation $\delta$ within an $L_\infty$ bound of $\epsilon_{AC} < \frac{w}{4}$, and the function $AC$, CAP holds.
\end{theorem}

The full proof is in the supplement (Section~\ref{sec:supp_proof}). The idea is that with perturbations of less than $\frac{w}{4}$, the midpoint between the farthest pairs of colors in the image will cleanly separate the clusters of pixels assigned to the original two colors. Thus, $AC$ will set all pixels to the left of the boundary to one color $c_1^*$ and all pixels to the right of the boundary to another color $c_2^*$ such that $c_1^* < c_2^*$. This is illustrated in Fig.~\ref{fig:proof} in the supplement (Section~\ref{sec:supp_proof}).

\begin{theorem}\label{thm:poc2}
Let $\mathcal{C_F}$ be any arbitrary fixed codebook. There exist images in MNIST$_G$ that, under $L_\infty$ perturbations of an $\epsilon$ bound of 1 with $FC_{\mathcal{C_F}}$ such that CAP does not hold.

\end{theorem}

The full proof is in the supplement (Section~\ref{sec:supp_proof}). The idea is that for any input images that contain a color lying at or within 0.5 of the boundary between two colors, with a perturbation $\delta$ with an $L_\infty$ bound of 1 we can arbitrarily set one pixel to the original codebook color while switching all other pixels to a different codebook color on the other side of the boundary, breaking CAP. This is illustrated in Fig.~\ref{fig:proof} in the supplement (Section~\ref{sec:supp_proof}).

\begin{theorem}\label{thm:poc3}
   For MNIST$_G$, we can guarantee that $M \circ S \circ AC$ achieves the same robust accuracy for an $L_\infty$ bound of $\epsilon < \frac{w}{4}$ as the natural accuracy of $M$ on MNIST$_D$. For a fixed codebook based defense, we cannot issue such a guarantee without further assumptions.
\end{theorem}
The full proof is in the supplement (Section~\ref{sec:supp_proof}). With CAP holding, we can reliably transform the output of $AC(x + \delta)$ back to the original MNIST$_D$ image that generated $x$ in the construction of MNIST$_G$, recovering the natural accuracy of $M$ on MNIST$_D$. However, without further knowledge, we do not know how robust a fixed codebook defense will be.

In summary, we show there exists datasets where an adaptive codebook strategy offers provably stronger robustness guarantees than a fixed codebook that uses the same codebook for every input image.

 

\section{Content-Adaptive Pixel Discretization}\label{sec:approach}


 The key idea behind \emph{content-adaptive pixel discretization} is to reconstruct the input 
 as faithfully as possible (high representation) with separable colors (high separability). In this way, we are able to cut down on the attack surface of an adversary. We select the 
 codebook colors on a 
 per-image basis rather than a 
 per-dataset basis as in prior work~\cite{chen2019towards}. This can be formulated as 
 the following, where $T_C$ takes each pixel in an image $x$ and sets it to the nearest color in the cluster palette $C$, and where $d$ 
 is the minimum distance between any two clusters:
 
 \begin{equation}\label{eq:joint}
    \begin{aligned}
        \argmin_{C} \; & \vert\vert T_C(x) - x\vert\vert_2 \\
         \textrm{s.t.} \mkern6mu & \forall_i\forall_j \hspace{0.2em} \vert\vert C_i - C_j \vert\vert_2 > d    \\
    \end{aligned}
\end{equation}

\subsection{Selecting an Adaptive Codebook}  We modify the ``Fast'' $k$-means color reduction process seen in Jalalpour \etal~\cite{jalalpour2019leveraging}, where $k$-means clustering is run on a thumbnail version of the image to find $k$ clusters to reconstruct the original image with. Specifically, we heuristically select the number of clusters $k$ adaptively based on the image content. This helps the algorithm adjust when some classes require more colors than others. We first change the initialization of clusters; rather than using random initialization, we bin 
the color space into $b \times b \times b$ cubes and initialize a cluster in the center of each cube that contains a color of any pixel in the image. This allows for good initial representation. Then, to enforce separability, we pare down the final cluster list by greedily iterating over the cluster list and accepting colors only if they are at least $d$ away from all other previously accepted colors.


\subsection{Optimization: Adaptive Gaussian Blurring}
One potential weakness of an adaptive discretization approach is that the adversary has some influence over the codebook selection. Thus, as an optimization, we empirically find that we can further improve the adaptive clustering assignments by applying adaptive, edge-aware Gaussian blurring. An adaptive blurring is motivated over vanilla Gaussian blurring because while a vanilla approach removes high frequency changes an adversary may attempt to add~\cite{raju2020blurnet,xu2017feature}, it also blurs important object edges. 
We thus propose an edge-aware adaptive blur, which attempts to preserve pertinent edges by selecting smaller kernels in areas with a high edge map response and selecting larger kernels in areas with roughly the same color to limit an adversary's ability to add misleading features.

\noindent\textbf{Computing Edges:}
We compute channel-wise edge response maps by taking the gradient magnitude of the standard Sobel~\cite{bradski2008learning} filters in the $x$ and $y$ directions. We then normalize it by dividing by 1140.4, the maximal attainable Sobel magnitude for the image range $[0..255]$.

We choose Sobel edge response maps over Canny Edge Detection~\cite{canny1986computational} because Sobel's finer-grained response maps allow further distinctions between weaker edges such as textures versus clear edges outlining the main object.
It is also more efficient for adversarial training.

 \begin{figure*}[t]
    \centering
    \includegraphics[width=0.85\textwidth]{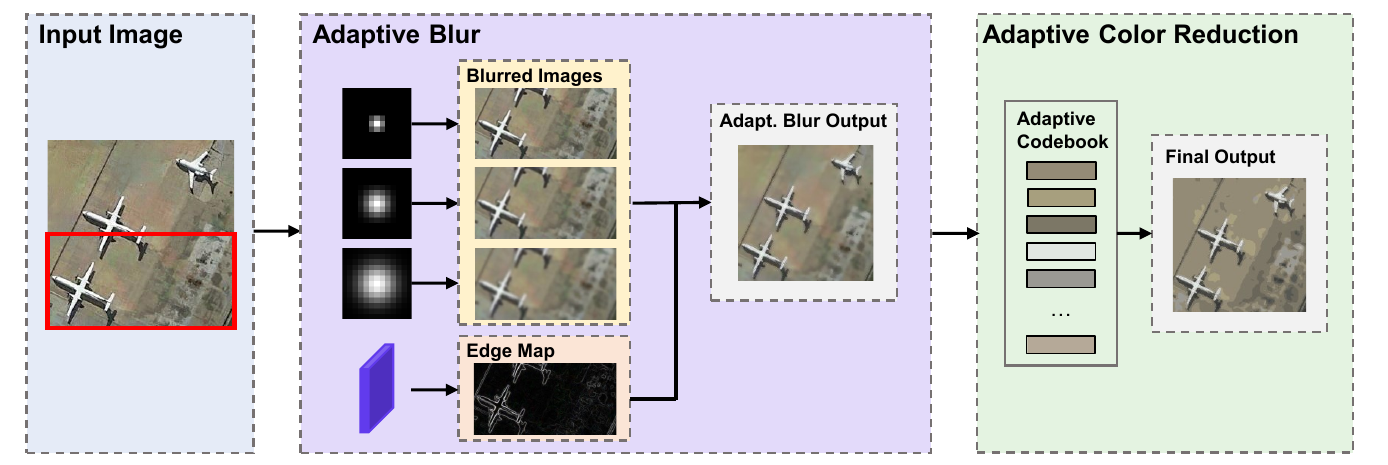}
\caption{\alg pipeline. \alg first applies adaptive blurring with different blur kernelds depending on the edge response map. \alg then applies content-adaptive pixel discretization to a codebook tailored to the per-image content.
} \label{fig:abc_pipeline}
\end{figure*} 

\noindent\textbf{Adaptive Kernel Thresholding:}\label{sec:threshold}
Once an edge response map has been computed, we then adaptively blur the image with different Gaussian kernels based on the edge responses. To preserve the edges, we want to apply small blur on strong edge pixels and 
large blur on pixels in the middle of an object.

Specifically, given a set of $m$ decreasing kernel sizes $z_1, z_2, ..., z_m$ to choose from and a set of increasing thresholds $t_1, t_2, ..., t_{m-1}$ we choose a blur level according to the edge response map. If the edge response at a pixel is $< t_1$, we apply $z_1$. Otherwise, let $t_i$ be the largest threshold the edge response is greater than. We then blur this pixel with $z_{i+1}$. Thus, in an area with no edges, we apply the most blur, and then for stronger and stronger edges, we apply less and less blur. This process is applied on each color channel separately. 

These thresholds are based on the selected threat model. We generally use the highest blur on edges less than or equal to the strength of edge increase allowed by the threat model; to have some protection on small edges that the adversary amplified, we double that threshold to set the range of edges that use the middle blur.
Note, one assumption this approach makes is that important natural edges are distinguishable in strength to edges an adversary can add.

 The final pipeline is shown in Fig.~\ref{fig:abc_pipeline}.

\section{Experiments}\label{sec:experiments}
We find experimentally that \alg significantly improves $L_2$ robustness over adversarial training baselines on complex datasets. \alg also improves the $L_\infty$ robustness on higher resolution datasets. We then analyze the behavior of \alg.

\subsection{Experimental Setup}\label{sec:exp_setup}
We describe the main points of our experimental setup here. Additional hyperparameters and details are included in the supplement (Section~\ref{sec:supp_extra_details}).

\noindent\textbf{Datasets and Models:}
We test on the CIFAR-10~\cite{krizhevsky2009learning}, RESISC45~\cite{cheng2017remote}, and Imagenette datasets~\cite{imagenette} datasets as they do not have a small, clearly defined set of dataset-wide colors that should be included in a potential codebook. Results on other datasets with clearer choices of colors that we would not expect our approach to significantly outperform a fixed codebook on (MNIST~\cite{lecun1998gradient}, Fashion-MNIST~\cite{xiao2017/online}, and GTSRB~\cite{Stallkamp2012}) are included in the supplement (Section~\ref{sec:supp_mnist}), where we generally find our approach is comparable to the fixed codebook approach. We use ATTA~\cite{zheng2020efficient} as an efficient adversarial training technique. For CIFAR-10, we use the Wide ResNet 34-10~\cite{zagoruyko2016wide} architecture commonly used in adversarial training techniques~\cite{madry2017towards,zhang2019theoretically,zheng2020efficient}. For RESISC45 and Imagenette, we use ResNet-34~\cite{he2016deep}.

\noindent\textbf{Attack Details:}
For PGD attack hyperparameters, we follow a similar setup to prior work~\cite{chen2019towards}. $L_2$ $\epsilon$ bounds are derived by setting them equal to what the $L_2$ distance would be for an attack that used the corresponding $L_\infty$ budget on each pixel. Exact details on hyperparameters and further attack details are included in the supplement (Section~\ref{sec:supp_attack_details}).

For adaptive attacks, we apply 
BPDA~\cite{athalye2018obfuscated}. 
For the adaptive discretization, we set $g(x)$ to the identity function. For 
adaptive blur, we differentiate exactly under the assumption that the choice of kernels will remain constant. For Chen~\etal~\cite{chen2019towards}, we use their proposed adaptive attack. We discuss alternatives in the supplement (Section~\ref{supp:adapt}).

We also test a black-box soft-label attack known as Square Attack~\cite{andriushchenko2020square} in the Supplement (Section~\ref{supp:black_box}) to provide a different style of attack and expose potential gradient masking issues~\cite{carlini2019evaluating,tramer2020adaptive}. We use 1000 iterations. For efficiency, we test on a 500 image subset. We find that these results do not significantly change the trends found in Section~\ref{sec:results}.


\subsection{Experimental Results}\label{sec:results}
Our main results on \alg are summarized in Table~\ref{tab:wb_table_complex}. Following Chen~\etal~\cite{chen2019towards}, we refer to models that were trained without the transform as ``pretrained'' and models that were trained with the transform as ``retrained''. We show results on four settings: 1) the pretrained ATTA baseline, 2) the pretrained ATTA baseline with fixed codebook discretization~\cite{chen2019towards}, 3) the retrained ATTA model with just adaptive discretization, and 4) the retrained ATTA model with the full \alg with adaptive discretization and blurring. \footnote{We found that other settings explored in Chen~\etal~\cite{chen2019towards} such as naturally pretrained models with discretization do not provide any significant improvement in any setting. We include results in the supplement (Section~\ref{supp:extra_training_modes}).}


\begin{table}[t]
\centering
\caption{White-box robustness results on more complex color datasets. We find that \alg improves $L_2$ and $L_\infty$ robustness on higher resolution datasets (RESISC45, Imagenette) and improves $L_2$ robustness on CIFAR-10. All models trained with ATTA~\cite{zheng2020efficient}.}\label{tab:wb_table_complex}
\begin{tabular}{@{}clccc@{}}
\toprule
\textbf{Dataset}                                                         & \multicolumn{1}{c}{\textbf{Model}} & \textbf{Accuracy} & \textbf{\begin{tabular}[c]{@{}c@{}}Robustness\\ ($L_2$ PGD)\end{tabular}} & \textbf{\begin{tabular}[c]{@{}c@{}}Robustness\\ ($L_\infty$ PGD)\end{tabular}} \\ \midrule
\multirow{4}{*}{CIFAR-10}                                                & Baseline                               & 85.43\%  & 20.15\%                                                                & 51.72\%                                                                  \\
                                                                         & Fixed Codebook~\cite{chen2019towards} (Pretrained)           & 84.73\%           & 26.42\%                                                                & \textbf{52.91\%}                                                         \\

                                                                         & Ours: AC (Retrained)  & 85.27\%           & 29.45\%                                                                                    & 50.14\% \\
                                                                                                                                                  & Ours: AC + AB (Retrained)              & 84.52\%           & \textbf{38.46\%}                                                       & 51.97\%                                                                  \\ \midrule

\multirow{4}{*}{RESISC45}                                                & Baseline                               & 82.00\%           & 12.09\%                                                                & 43.89\%                                                                  \\
                                                                                                & Fixed Codebook~\cite{chen2019towards} (Pretrained)           & 81.89\%           & 16.36\%                                                                & 46.27\%                                                                  \\
                                                                        
                                                                         & Ours: AC (Retrained) & 83.44\%           & 30.80\%                                                                                  & 45.24\% \\
                                                                          & Ours: AC + AB (Retrained)             & 88.24\%  & \textbf{52.44\%}                                                       & \textbf{53.33\%}                                                         \\ \midrule
\multirow{4}{*}{Imagenette}                                              & Baseline                               & 93.58\%           & 65.50\%                                                                & 75.97\%                                                                  \\
                                                                        
                                                                         & Fixed Codebook~\cite{chen2019towards} (Pretrained)           & 93.45\%           & 70.83\%                                                                & 78.70\%                                                                  \\
                                                                                              
                                                                         & Ours: AC (Retrained) & 94.75\%           & 71.62\%                                                                                  & 79.90\% \\
                                                                         & Ours: AC + AB (Retrained)             & 95.34\% & \textbf{72.28\%}                                                       & \textbf{80.71\%}                                   \\ \bottomrule
\end{tabular}
\vspace{-0.1in}
\end{table}

\noindent\textbf{$L_2$ Robustness:}
We find that robustness against $L_2$ attacks improves with adaptive codebooks for all three datasets (29.45\%, 30.80\%, and 71.62\% for CIFAR-10, RESISC45, and Imagenette respectively) over the pretrained ATTA and pretrained ATTA with fixed codebook~\cite{chen2019towards} baselines.

We then find that the addition of adaptive blurring further raises the adversarial robustness against $L_2$ attacks for all datasets, with CIFAR-10 improving from 29.45\% to 38.46\%, RESISC45 improving from 30.80\% to 52.44\%, and Imagenette improving from 71.62\% to 72.28\%. We think the dramatic improvement in RESISC45 is likely due to its higher resolution  and detail as well as its tendency to include images with large areas of similarly clustered colors where blur is more likely to be effective. More analysis is included in Section~\ref{sec:analysis}.

\noindent\textbf{$L_\infty$ Robustness:}
With just the adaptive codebook, we find that the adversarial robustness against $L_\infty$ attacks is comparable to the pretrained ATTA and pretrained ATTA with fixed codebook~\cite{chen2019towards} baselines. However, we find that on the higher resolution datasets that the addition of adaptive blurring helps improve the $L_\infty$ robustness over using just the adaptive codebook, with RESISC45 increasing from 45.24\% to 53.33\% Imagenette increasing from 79.90\% on just the adaptive codebook to 80.71\%. As with $L_2$ robustness, these datasets benefit from having higher resolution and thus more ability to blur and discretize aggressively without removing essential features of the original image. In contrast, we expect the lack of improvement in CIFAR-10 to be due to a lack of detail in the original images, limiting our ability to clean up noise while still preserving key features.

\noindent\textbf{Natural Accuracy:} While not our focus, we find that the natural accuracy improves with adaptive codebooks on RESISC45 and Imagenette with comparable natural accuracy on CIFAR-10 against the pretrained ATTA and pretrained ATTA with fixed codebook~\cite{chen2019towards} baselines. We see further 4.8\% improvement to 88.24\% on RESISC45 when adding in the adaptive blur. Like with $L_2$ and $L_\infty$ robustness, RESISC45 and Imagenette likely benefited from its significantly larger resolution, with RESISC45 also containing images with large areas of similarly clustered colors where blurring is likely to be more effective.

\subsection{Discussion and Limitations}\label{sec:analysis} 
We now explore how the $L_2$ attack space is restricted with an example from RESISC45 and limitations. In the supplement, we also include ablations (Section~\ref{supp:ablate}) on just adaptive blurring or adaptive discretization and an analysis of attack perceptibility (Section~\ref{sec:supp_percept}).

\noindent\textbf{$L_2$ Results:}
The way we set the $L_2$ bounds to include the $L_\infty$ space means that the space of $L_\infty$ is a subset of the $L_2$ attack space. Thus, $L_2$ robustness will be less than the $L_\infty$ robustness. It is then interesting that the robustness increases more dramatically for $L_2$ than $L_\infty$. This is especially apparent for RESISC45, where the $L_2$ robustness is nearly brought to the $L_\infty$ robustness (52.44\% vs. 53.33\%). This means that the added space included in the $L_2$ ball but not the $L_\infty$ ball barely adds any additional attacks. 

\begin{figure}[t]
    \centering
    {\makebox[1.2in]{\includegraphics[width=0.8in]{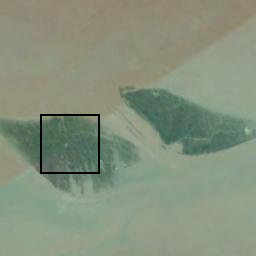}}}%
    \hfil
    {\makebox[1.2in]{\includegraphics[width=0.8in]{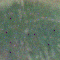}}}%
    \hfil
    {\makebox[1.2in]{\includegraphics[width=0.8in]{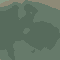}}}%
    \caption{$L_2$ attack example on RESISC45 ATTA baseline. $L_2$ attacks consisting of isolated pixels with large perturbations no longer work as well. \textbf{Left}: an original image of an island from RESISC45 with ROI box. \textbf{Center}: adversarial version of ROI, with visible black / pink perturbations. \textbf{Right}: \alg version that removed the black / pink perturbations. }
    \label{fig:explainl2}
    \vspace{-0.1in}
\end{figure}

We understand this dramatic increase when adding blur to be a consequence of how blurring and adaptive discretization can work together. In particular, if the adaptive blur can eliminate isolated pixel changes in non-edge regions well enough that adaptive discretization sends it to the correct cluster, we can undo the effects of that perturbed pixel. As an example, we show in Fig.~\ref{fig:explainl2} a perturbed $L_2$ attack of an island from the RESISC45 against the pretrained ATTA baseline. We find that the attack left some isolated but noticeable black and pink perturbations. However, when applying \alg, these individual changes disappeared, showing the types of attacks that are no longer available. This kind of attack surface reduction is also likely available in CIFAR-10 and Imagenette but to lesser degrees.

\noindent\textbf{Limitations:} On other datasets where there are a limited set of dataset-wide colors (Supplement Section~\ref{sec:supp_mnist}), our approach may not provide any improvement over a fixed codebook. \alg also may struggle with datasets that are too low in resolution and detail. Nevertheless, we find that \alg can provide significant robustness improvements on datasets when a fixed codebook fails. Thus, \alg and its adaptive discretization should be considered a valuable part of the repertoire of preprocessing defense techniques to try out for improving adversarial robustness of image classification pipelines.


\section{Conclusion}
We propose \alg, a content-adaptive pixel discretization defense to improve model robustness. We first formally proved that adaptive codebooks have stronger robustness guarantees than fixed codebooks on MNIST$_G$. We then found empirically that on complex datasets with fewer dataset-wide color trends, our per-image adaptive approach can improve the $L_2$ and $L_\infty$ robustness where a fixed codebook strategy fails to do so. Thus, integrating \alg as a potential defense strategy can help extend the number of datasets for which robustness can be increased with discretization-based approaches.

\bibliographystyle{IEEEtran}
\bibliography{references}

\appendix

\section{Proofs}\label{sec:supp_proof}
\setcounter{theorem}{0}
\begin{theorem}
Given any arbitrary image $x \in $ MNIST$_G$, any arbitrary integer perturbation $\delta$ within an $L_\infty$ bound of $\epsilon_{AC} < \frac{w}{4}$, and the function $AC$, CAP holds.
\end{theorem}
\begin{proof}
By construction, $AC$ generates output images of two colors. Let $x'$ refer to the arbitrarily given input image $x' = x + \delta$ where $x \in$ MNIST$_G$ and $\delta$ is an integer perturbation such that $\vert\vert\delta\vert\vert_\infty \leq \frac{w}{4}$. Let $c_1$, $c_2$ refer to the colors in the original, unperturbed image $x$ where without loss of generality $c_1$ refers to the smaller color of the two. Let $P_1 = \{(i, j) : x[i, j] = c_1 \}$ and $P_2 = \{(i, j) : x[i, j] = c_2\}$ refer to the two disjoint sets of pixels by color value in $x$. Then, in $x'$ with perturbations of $\epsilon < \frac{w}{4}$, we know that the value range of pixels in $P_1$ is $(c_1 - \frac{w}{4}, c_1 + \frac{w}{4})$ and the value range of pixels in $P_2$ is $(c_2 - \frac{w}{4}, c_2 + \frac{w}{4})$. By definition of MNIST$_G$, we know that $c_2 - c_1 \geq w$. This means that the range for $c_{mid}$ is $[c_1 + \frac{w}{4}, c_2 - \frac{w}{4}]$. Because $c_{mid}$ must lie between the ranges that pixels in $P_1$ and $P_2$ can be, all pixels in $P_1$ will be assigned to the same pixel as each other in $AC(x')$ and all pixels in $P_2$ will be assigned to the same pixel as each other in $AC(x')$. So, the Cluster Assignment Property and thus Theorem~\ref{thm:poc1} holds.
\end{proof}

\begin{theorem}
Let $\mathcal{C_F}$ be any arbitrary fixed codebook. There exist images in MNIST$_G$ that, under $L_\infty$ perturbations of an $\epsilon$ bound of 1 with $FC_{\mathcal{C_F}}$ such that CAP does not hold.
\end{theorem}

\begin{proof}
Let $c_a$ and $c_b$ refer to any arbitrary pair of adjacent codebook colors in $\mathcal{C}_F$. Without loss of generality, let $c_a$ refer to the smaller of the two colors. Let $c_{mid} = \lfloor(c_a + c_b) / 2\rfloor$. Now, suppose we encounter an input example $x$ from MNIST${}_G$ that consists of the two colors $c_1 = c_{mid}$ and $c_2 = c_{mid} + w$ (or case 2, if $c_{mid} + w > 255$, select an input with $c_1 = c_{mid} - w$ and $c_2 = c_{mid}$). Let $P = \{(i, j) : x[i, j] = c_{mid}\}$ be the set of pixels originally assigned to $c_{mid}$. Under a perturbation bound of $\epsilon = 1$, an adversary can violate the CAP with an attacked image $x' = x + \delta$ where $\vert\vert \delta \vert\vert_\infty \leq \epsilon$ by leaving one pixel in $P$ untouched and adding 1 to the rest of the pixels in $P$. Then, the pixels that had $\epsilon$ added to them are guaranteed to cross the boundary from closer to $c_a$ to closer to $c_b$. This sets the one pixel that was untouched to $c_a$ while all other pixels from $P$ are set to $c_b$, violating the CAP.



Both cases hold, so Theorem~\ref{thm:poc2} holds. 

\end{proof}

\begin{theorem}
   For MNIST$_G$, we can guarantee that $M \circ S \circ AC$ achieves the same robust accuracy for an $L_\infty$ bound of $\epsilon < \frac{w}{4}$ as the natural accuracy of $M$ on MNIST$_D$. For a fixed codebook based defense, we cannot issue such a guarantee without further assumptions.
\end{theorem}

\begin{proof}
Theorem~\ref{thm:poc1} implies that we can guarantee robustness for an adaptive codebook defended model $M \circ S \circ AC$, where $S$ takes the CAP satisfying output from $AC$, discovers the two color values being used, and remaps the lower color to 0 and the higher color to 255. We then know that the output of $S(AC(x + \delta))$ where $x \in $ MNIST$_G$ and $\delta$ is an integer perturbation such that $\vert\vert\delta\vert\vert_\infty < \frac{w}{4}$ will be the same as the original image from MNIST$_D$ that $x$ was constructed from. Thus, our robust accuracy on $M \circ S \circ AC$ on MNIST$_G$ is equal to the natural accuracy of $M$ on MNIST$_D$.

Conversely, Theorem~\ref{thm:poc2} implies that we can not guarantee any robustness for a fixed codebook-based defense without assuming anything further about $M$ or $\mathcal{C_F}$. Without CAP holding, we do not know a priori what $M(x + \delta)$ will be or what the original MNIST$_D$ image was. Thus, Theorem~\ref{thm:poc3} holds.
\end{proof}

\noindent\textbf{Choice of $w$:} Note that the theorems hold even for $w$ in [1..4]. However, in the main paper, we set $w \geq 5$ because $1 \leq w \leq 4$ are essentially degenerate cases where the robustness guarantee is also 0, and thus just as bad as the fixed codebook. With a $w \geq 5$, the adaptive codebook at least can guarantee a robustness of perturbations up to and including 1, the smallest useful bound in this case given we are working with integer images.

\begin{figure}[t]
     \centering
       {\makebox[4.5in]{\includegraphics[width=2.5in]{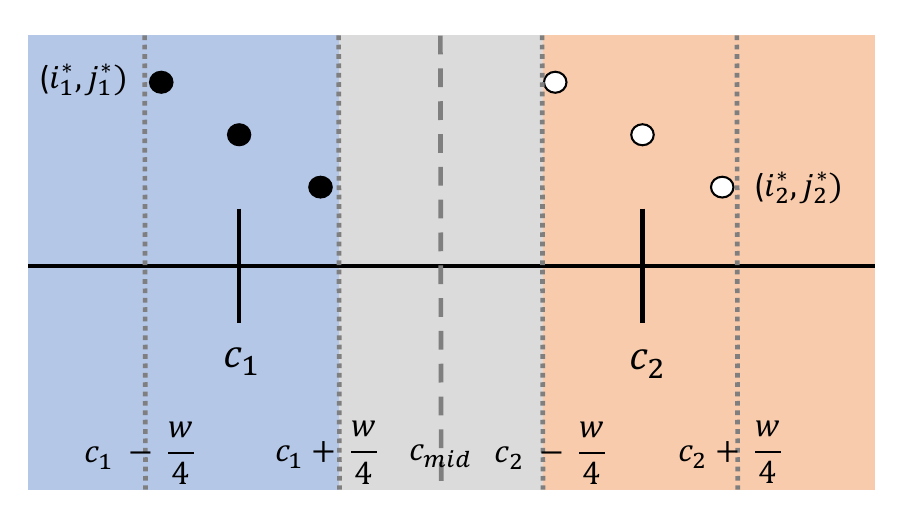}\hfill\includegraphics[width=2.5in]{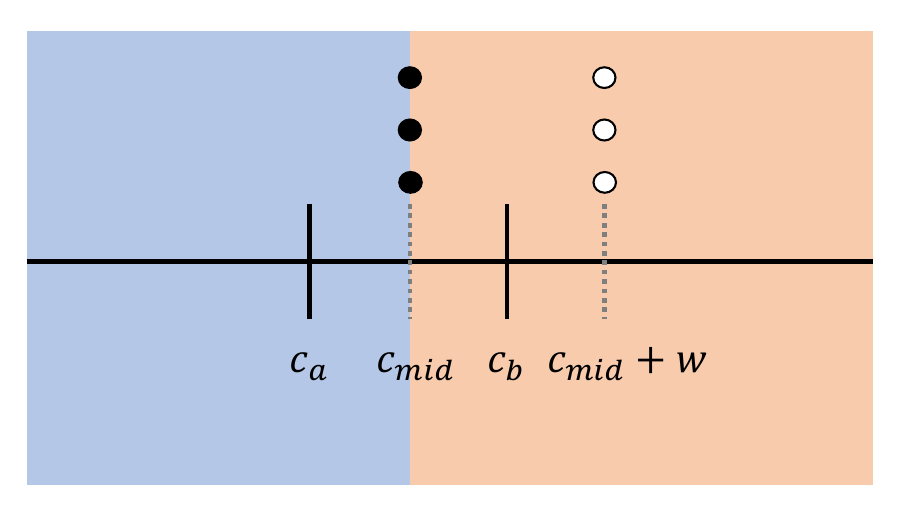}}}%
       
    \hfill
    \caption{Proof visualization. Left: Conversely, in an adaptive codebook, we can tolerate perturbations of $\frac{w}{4}$. The points originally assigned to $c_1$ are bounded in the blue region and the points originally assigned to $c_2$ are bounded in the orange. The calculated value for $c_{mid}$ is guaranteed to fall within the gray region between the perturbed clusters of points. Right: A fixed codebook with colors $c_a$ and $c_b$ cannot tolerate perturbations to an input image with colors $c_{mid}$ and $c_{mid} + w$, since all black points at $c_{mid}$ could arbitrarily be assigned either to the blue or orange space. }
    \label{fig:proof}
\end{figure}

\section{Additional Experimental Details}\label{sec:supp_extra_details}
This section includes additional details on both the datasets used in the main paper and also those with additional results in the supplement.

\noindent\textbf{Datasets and Models:}
For MNIST and Fashion-MNIST, we use the MNIST architecture used in prior work~\cite{madry2017towards}. For GTSRB, we use the Wide ResNet 34-10~\cite{zagoruyko2016wide} architecture commonly used in adversarial training techniques~\cite{madry2017towards,zhang2019theoretically,zheng2020efficient}.

For GTSRB, similar to prior work we crop and scale all images to 32 $\times$ 32~\cite{chen2019towards,roadsigns17}. Our specific process entails cropping the image at the given annotation region-of-interest (ROI) coordinates included with the data download, padding the crop to a square and resizing the square to $32 \times 32$. This way, the crops can be used in a consistent $32 \times 32$ size without changing the distortion or aspect ratio of the images. Then, like Chen~\etal~\cite{chen2019towards}, we remove all images with an average intensity $<$ 50 from GTSRB. 

\noindent\textbf{Attack Details:}\label{sec:supp_attack_details}
For our $L_\infty$ PGD-based attacks 
our setup is similar to prior work~\cite{chen2019towards}. Specifically, for MNIST, we use $\epsilon = 0.3$ and 100 steps of size $0.01$. For Fashion-MNIST, we use $\epsilon = 0.1$ and 100 steps of size $0.01$. For CIFAR-10, we test with $\epsilon = 0.031$ and 40 steps of size 0.007. For GTSRB and RESISC45, we use $\epsilon = 0.031$ and 40 steps of size 0.007. For Imagenette, we test with $\epsilon = 4/255$ and 10 steps of size 1/255.

For $L_2$ attacks, we adjust the $\epsilon$ values by setting them equal to what the $L_2$ distance would be for an attack that used the corresponding $L_\infty$ budget on each pixel. For example, for CIFAR-10, perturbing each value by $\epsilon = 8/255$ would have a distortion of $8/255 \cdot \text{sqrt}(32\cdot32\cdot3) = 1.74$, so we set this $L_2$ $\epsilon$ to 1.74. For MNIST, Fashion-MNIST, CIFAR-10, GTSRB, RESISC45, and Imagenette, we use step sizes of 0.1, 0.5, 0.05, 0.05, 0.35, and 0.85 respectively.

We also modify PGD to return the attack from the step with the highest loss rather than simply returning the attack from the last step. This helps improve the attack success rate on models with \alg without noticeably impacting the baseline evaluation. All future PGD references assume this change.

\noindent\textbf{Training Details:}
For MNIST and Fashion-MNIST, we train natural models for 40 epochs with a learning rate of 0.01 for 35 epochs and 0.001 for the remaining. We use a batch size of 128, weight decay of 2e-4, and 0.9. we use settings as in~\cite{zheng2020efficient} with 100 epochs and a learning rate division of 10 at epochs 55, 75, and 90. For the ATTA-based models, we use a batch size of 64, train for 60 epochs without resetting the perturbations, momentum 0.9, starting learning rate of 0.1 $\cdot$ 1/64 and weight decay of 2e-4 $\cdot$ 64, and we divide the learning rate by 10 after 55 epochs\footnote{We confirmed through correspondence with the original authors that there was an additional batch size division in the code. Adjusting the values as we have replicates the original code's behavior for a learning rate input of 0.1 and weight decay of 2e-4.}. The training time attacks were PGD-40 with a step size of 0.01 and $\epsilon = 0.3$ for MNIST and PGD-20 with a step size of 0.01 and $\epsilon = 0.1$ for Fashion-MNIST, following prior work~\cite{chen2019towards}. 

For CIFAR-10 and GTSRB, we train natural models the same as MNIST but with the learning rate 10x as high (starting at 0.1). For ATTA-based models, we use 38 epochs, a perturbation reset of 10 epochs, a starting learning rate of 0.1 $\cdot$ 1/64 with a division by 10 at epochs 30 and 35, and 2e-4 $\cdot$ 64 for weight decay, following prior work~\cite{zheng2020efficient} (see footnote 3). The training time attacks were PGD-7 with a step size of 2/255 = 0.007 and $\epsilon$ of 8/255 = 0.031.

For RESISC45, we use the same settings as MNIST for natural training. For ATTA-based models, we use 40 epochs, a perturbation reset of 10 epochs, a starting learning rate of 0.1 $\cdot$ 1/64 with a division by 10 at epochs 30 and 35, and 2e-4 $\cdot$ 64 for weight decay. The training time attacks were PGD-10 attacks with a step size of 2/255 = 0.007 and an $\epsilon$ of 8/255 = 0.031.

For Imagenette, we use the same settings as MNIST and RESISC45 for natural training. For ATTA-based models, we use the same settings as RESISC 45 except the training time attacks were PGD-10 attacks with a step size of 1/255 = 0.004 and an $\epsilon$ of 4/255 = 0.0157.

Additionally, for ATTA experiments on GTSRB, RESISC45, and Imagenette, we save out intermediate results rather than caching the entire dataset in memory as the original code does due to memory concerns. Following previous work~\cite{madry2017towards,zheng2020efficient}, we apply random horizontal flipping and random cropping data augmentation. For CIFAR-10, as in prior work~\cite{madry2017towards,zheng2020efficient}, we apply padding of 4 pixels on each side and randomly crop back to $32 \times 32$. For RESISC45 and Imagenette, we apply padding of 2 pixels on each side and randomly crop back to the original resolution.

\noindent\textbf{\alg Transformation Details:}
For adaptive blurring, on the 32 $\times 32$ datasets (CIFAR-10 and GTSRB), we use the smallest three possible kernel widths of \{5, 3, 1\}. A kernel of 5 is still a large amount of blur however, covering over 15\% of the image length. MNIST and Fashion-MNIST are even smaller, so we use \{3, 1, 1\}. For RESISC45, we use kernels of size \{13, 7, 3\}, which is about 5\%, 2.5\% and 1\% of the 256 image length. For Imagenette, we use \{11, 5, 3\} to maintain similar ratios as RESISC45. The kernels use the default standard deviation from OpenCV.

We set the kernel thresholds equal to \{20, 40\} for each dataset except for Imagenette, which we reduce to \{10, 20\} given the smaller epsilon bounds. The \{20, 40\} thresholds 
ensure that the strength of edges an adversary could apply under the $L_\infty$ bounds of CIFAR-10, GTSRB, and RESISC45 would be below these values (and similarly with \{10, 20\} for Imagenette).

For our color reduction process, we bin the color space up into cubes of length $b = 16$ for cluster initialization. We set 32 $\times$ 32 to be the thumbnail size for Fast $k$-means~\cite{jalalpour2019leveraging}. Then, when paring down the cluster list, we enforce a minimum distance $d$ equal to 3x the size of $\epsilon$. This way, should an adversary attempt to take two pixels at the center of two clusters and switch them, there are two radii of $\epsilon$ plus and $\epsilon$ length of buffer between them, to minimize the number of pixels where the adversary could choose which cluster they belong to.\footnote{Note that for MNIST, with an $\epsilon$ of 0.3, this works out to be 230 / 255. Thus, we tweak the coalescing algorithm to avoid choosing a cluster that would automatically rule out the ability to pick a second cluster. If no such pair of clusters obey this distance, we simply choose the farthest two apart.}

For $k$-means color reduction, as in Jalapour \etal~\cite{jalalpour2019leveraging}, we approximate the choice of colors on a smaller thumbnail image before assigning the original image's colors to that palette. We downsize the image to a resolution of $32 \times 32$ for the thumbnail version. We use OpenCV's area interpolation method. We run $k$-means clustering for 20 iterations.

\noindent\textbf{Fixed Codebook Details:}
For experiments involving a fixed codebook~\cite{chen2019towards}, on common datasets we take the codebook settings from the most robust settings from the original paper. The specific parameters are $k$, the number of codewords, and $r$, a distance parameter. For MNIST and Fashion-MNIST, we use $k=2$ and $r=0.6$ (in [0-1] image range). For CIFAR-10 and RESISC45, we use $k=300$ and $r=16$ (in [0-255] image range), and for GTSRB and Imagenette we use $k=500$ and $r=16$.

For the adaptive attack settings, as in the original paper~\cite{chen2019towards}, we use a value of 10 for their attack parameter $\alpha$ on MNIST and Fashion-MNIST and a value of 0.1 for other datasets.

\section{Main Results on MNIST, Fashion-MNIST, and GTSRB}\label{sec:supp_mnist}

Results for MNIST, Fashion-MNIST, and GTSRB are shown in Table~\ref{tab:wb_table_simple} on the same defense types as Table~\ref{tab:wb_table_complex} in addition to the pretrained ATTA with adaptive color reduction and adaptive blur setting. These three datasets differ from those in the main paper as the difference between a fixed codebook and an adaptive codebook could feasily be minimal, as they generally have distinct colors that appear for large portions of the datasets.

\begin{table}[t]
\centering
\caption{White-box robustness results on simpler grayscale datasets. Adding \alg at test time improves  $L_2$ and $L_\infty$ robustness over vanilla ATTA, while retraining improves $L_\infty$ robustness on Fashion-MNIST. We show in the supplement (Section~\ref{supp:black_box}) that Chen~\etal's $L_2$ robustness is likely overstated; once adjusted, Chen~\etal performs comparably to \alg. }\label{tab:wb_table_simple}
\begin{tabular}{@{}clccc@{}}
\toprule
\textbf{Dataset}                                                         & \multicolumn{1}{c}{\textbf{Model}} & \textbf{Accuracy} & \textbf{\begin{tabular}[c]{@{}c@{}}Robustness\\ ($L_2$ PGD)\end{tabular}} & \textbf{\begin{tabular}[c]{@{}c@{}}Robustness\\ ($L_\infty$ PGD)\end{tabular}} \\ \midrule
\multirow{6}{*}{MNIST}                                                   & Baseline                               & 96.53\%           & 0.00\%                                                                 & 87.47\%                                                                  \\
                                                                         
                                                                         & Fixed Codebook~\cite{chen2019towards} (Pretrained)           & 96.30\%           & \textbf{28.75\%}                                                       & \textbf{93.85\%}                                                         \\
                                                                          & Ours: AC (Pretrained) & 96.28\%           &                             8.67\%                                                          & 93.21\% \\  
                                                & Ours: AC (Retrained) & 98.51\%           & 0.00\%                                                                                      & 91.73\%\\    & Ours: AC + AB (Pretrained)             & 96.29\%           & 13.89\%                                                                & 92.76\%                                                                  \\                      & Ours: AC + AB (Retrained)            & 98.43\%  & 0.00\%                                                                 & 91.00\%                                                                  \\
                                                                        \midrule
\multirow{6}{*}{\begin{tabular}[c]{@{}c@{}}Fashion-\\ MNIST\end{tabular}} & Baseline                               & 81.74\%           & 13.82\%                                                                & 69.80\%                                                                  \\
                                                                                              & Fixed Codebook~\cite{chen2019towards} (Pretrained)           & 78.95\%           & \textbf{43.41\%}                                                       & 66.69\%                                                                  \\ 
                         & Ours: AC (Pretrained) & 81.53\%           &                                26.69\%                                                       & 71.15\% \\                      & Ours: AC (Retrained) & 82.22\%           & 22.06\%                                                                                    & \textbf{71.88\%}                                                                                       \\ 
                                             & Ours: AC + AB, (Pretrained)             & 81.27\%           & 30.15\%                                                                & 70.61\%                                                                  \\& Ours: AC + AB (Retrained)             & 82.06\%  & 23.37\%                                                                & 71.54\%                                                         \\ \midrule
\multirow{6}{*}{GTSRB}                                                   & Baseline                               & 92.47\%  & 57.28\%                                                                & 78.24\%                                                                  \\
                                                                                            & Fixed Codebook~\cite{chen2019towards} (Pretrained)           & 92.31\%           & \textbf{59.82\%}                                                       & \textbf{81.16\%}                                                         \\
                           & Ours: AC (Pretrained) & 91.98\%           &          58.80\%                                                                             & 77.05\% \\                      & Ours: AC (Retrained) & 91.86\%           & 58.7\%                                                                                    & 76.45\%   \\     & Ours: AC + AB (Pretrained)             & 89.22\%           & 57.20\%                                                                & 71.93\%                                                                  \\                      & Ours: AC + AB (Retrained)             & 90.32\%           & 57.71\%                                                                & 73.25\%                                                                  \\ 
                                                                        \bottomrule
\end{tabular}
\end{table}

\noindent\textbf{$L_2$ Analysis:}
For MNIST and Fashion-MNIST, the fixed codebook approach~\cite{chen2019towards} appears to perform better than our \alg settings on these white-box settings, but we find in Section~\ref{supp:black_box} that Square Attack greatly reduces the robustness for the fixed codebook setting (28.75\% to 13.0\% on MNIST, 43.41\% to 27.8\% on Fashion-MNIST on 500 samples, which would be lower than the corresponding results for the pretrained ATTA with adaptive codebook and blurring setting). This suggests that the adaptive attack from Chen~\etal~\cite{chen2019towards} is overestimating the robustness of the fixed codebook approach. Note that there were no other cases in which our 500 image Square Attack test performed notably better than PGD (see Section~\ref{supp:black_box}).

We additionally add that for MNIST and Fashion-MNIST, we find that the pretrained ATTA with \alg generally is better than retrained ATTA with \alg settings. We suspect this is due to the larger perturbation bounds giving the adversary too much influence on the color palette to learn a good boundary for $L_2$ attacks.

On GTSRB, we find that the $L_2$ robustness does not significantly improve with \alg on GTSRB, although it is comparable to the baseline settings. We argue in the Section~\ref{sec:supp_gtsrb} that this is due to the reliance on small, high frequency differences in a few key areas to distinguish between multiple classes and show an example of an image with key distinguishing features removed. 

\noindent\textbf{$L_\infty$ Analysis:} On MNIST and Fashion-MNIST, we find that \alg improves the $L_\infty$ robustness over the pretrained ATTA model, with it also outperforming the fixed codebook approach for Fashion-MNIST. n the simple MNIST dataset where black and white discretization works extremely well, \alg performs comparably to Chen~\etal~\cite{chen2019towards}. 

On GTSRB, we find that the robustness gets slightly worse with \alg, although comparable to the baselines. We explain this by showing a similar trend to the $L_2$ results in supplement (Section~\ref{sec:supp_gtsrb}).

Unlike RESISC45 and Imagenette, we generally find that adding blur to these datasets does not help. This is likely because these datasets are too low resolution and too low detail for the blur to serve its purpose well without destroying the underlying natural features.

\noindent\textbf{Natural Accuracy:}
The natural accuracy is comparable in all settings with around 97\% on MNIST, 80\% on Fashion-MNIST, and 91\% on GTSRB. This is perhaps unsurprising as the the digits, clothing articles, and central traffic sign features are already well clustered, generally speaking.

\section{Alternative Adaptive Attacks}\label{supp:adapt}
We explored a few other adaptive white-box attack strategies but found them to be no better than using the identity function for color reduction and differentiating through the adaptive blur with the assumption the choice of kernel remains static, as described in the main paper. The first alternative attack tried adding an additional term to the attack objective function to maximize the average Sobel map response to encourage the adversary to add edges. We note that due to edge thresholding in the adaptive kernel selection (Section~\ref{sec:threshold}) the attacker is not able to add very much to the edge response within small $L_\infty$ ball limits, which limits the ability to attack the image by adding lots of strong edges. The second alternative attack tried to add a Non Printability Score~\cite{glasses} to encourage the attack to use very few colors. We tried using the set of colors used in~\cite{roadsigns17} and from the palette generated by $k$-means on the original input image, but neither improved the attack success rate. Third, for the color reduction stage, we tried using the same approximation as Chen \etal~\cite{chen2019towards} for the currently selected cluster at each step. However, initial testing found this to be less effective than the identity function, so we focus our evaluations with the identity function at the color reduction stage. 

\section{Alternative Combinations of Training + Transforms}\label{supp:extra_training_modes}
We include results on other training modes such as natural training, adding a transform to a naturally trained model, etc. in Table~\ref{tab:supp_c}. We generally find these settings to be inferior to the ones in the main paper. We find that on CIFAR-10 retrained ATTA models with fixed codebooks~\cite{chen2019towards} can perform slightly better on $L_\infty$ attacks, but not by much (53.18\% on retrained ATTA model with fixed codebook and 52.91\% on pretrained ATTA model with fixed codebook). 

We also observe that unless we expect the data to have well split colors already (i.e. MNIST, Fashion-MNIST, GTSRB), models where the transform is merely added on at test time perform inferior to those that are retrained on the transform. This is consistent with the results from the main paper too. We also observe that adversarial training seems to be required for optimal performance.

\section{Black-box Evaluation}\label{supp:black_box}
We include results on 500 samples of Square Attack~\cite{andriushchenko2020square} with 1000 iterations in Tables~\ref{tab:square1}, \ref{tab:square2}, and \ref{tab:square3} for the settings presented in Tables~\ref{tab:wb_table_complex} and~\ref{tab:wb_table_simple} in the main paper and in the additional combinations presented in Section~\ref{supp:extra_training_modes} in the Supplement.

We do not see any large reductions on models with \alg, suggesting that our \alg adaptive attacks are reasonable. We observe that on CIFAR-10, GTSRB, RESISC45, and Imagenette, Square Attack never improves the attack on a model with \alg involved in any capacity except for RESISC45 adversarially trained with \alg, but the drop is less than 3\% and within the noise of the 500 samples for Square Attack, suggesting that the adaptive attacks on \alg are good. The Square Attack numbers in several cases are also notably lower than the PGD numbers (such as the Natural + EF at Test Time and ATTA + EF at Test time settings for RESISC45 and Imagenette), suggesting that the adaptive attacks worked particularly well in these cases.

There are however some notable decreases on models with a fixed codebook~\cite{chen2019towards}, for which we used the original paper's adaptive attack. Notable decreases include MNIST on pretrained ATTA with fixed codebook ($L_2$ decreased from 28.75\% to 13.0\%) and Fashion-MNIST on pretrained ATTA with fixed codebook ($L_2$ decreased from 43.41\% to 27.8\%), as this decreases these models performance as originally seen from the PGD attacks. These results suggest that the adaptive attack for Chen~\etal~\cite{chen2019towards} could be further improved for a truer robustness estimate.
\begin{table}
\centering
\caption{White-box robustness results on alternative training modes. Generally, these modes achieve inferior performance to the versions reported in the main paper.}\label{tab:supp_c}
\begin{tabular}{@{}clccc@{}}
\toprule
\textbf{Dataset}                                                          & \multicolumn{1}{c}{\textbf{Model}} & \textbf{Accuracy} & \textbf{\begin{tabular}[c]{@{}c@{}}Robustness\\ ($L_2$ PGD)\end{tabular}} & \textbf{\begin{tabular}[c]{@{}c@{}}Robustness\\ ($L_\infty$ PGD)\end{tabular}} \\ \midrule
\multirow{6}{*}{MNIST}                                                    & F. Codebook~\cite{chen2019towards}, Nat. (Pretrain.)        & 99.07\%           & 0.03\%                                                                 & 77.37\%                                                                  \\
                         & F. Codebook~\cite{chen2019towards}, Nat. (Retrain.)        & 99.17\%           & 0.04\%                                                                 & 79.79\%                                                                  \\     
                         
                                                                                                   & F. Codebook~\cite{chen2019towards}, ATTA (Retrain.)           & 98.47\%           & 0.17\%                                                                 & 91.60\%                                                                  \\ 
                         
                         & Ours: AC + AB, Nat. (Pretrain.)          & 99.05\%           & 0.00\%                                                                 & 74.99\%                                                                  \\
                                                                   
                                                                          & Ours: AC + AB, Nat. (Retrain.)          & 99.06\%           & 0.00\%                                                                 & 77.49\%                                                                  \\

                                                                          & Ours: AC + AB, ATTA (Pretrain.)             & 96.29\%           & 13.89\%                                                                & 92.76\%                                                                  \\\midrule
\multirow{6}{*}{\begin{tabular}[c]{@{}c@{}}Fashion-\\ MNIST\end{tabular}} 
                                  
                          & F. Codebook~\cite{chen2019towards}, Nat. (Pretrain.)        & 81.96\%           & 7.90\%                                                                 & 40.56\%                                                                  \\
                          & F. Codebook~\cite{chen2019towards}, Nat. (Retrain.)        & 88.38\%           & 10.67\%                                                                & 47.63\%                                                                  \\
                          & F. Codebook~\cite{chen2019towards}, ATTA (Retrain.)           & 80.97\%           & 40.03\%                                                                & 69.23\%                                                                  \\ 
                          & Ours: AC + AB, Nat. (Pretrain.)          & 89.42\%           & 3.95\%                                                                 & 29.42\%                                                                  \\
                                                                        
                                                                          & Ours: AC + AB, Nat. (Retrain.)          & 90.54\%           & 5.86\%                                                                 & 32.79\%                                                                  \\
                                                                          
                                                                           & Ours: AC + AB, ATTA (Pretrain.)             & 81.27\%           & 30.15\%                                                                & 70.61\%                                                                  \\
                                                   \midrule
\multirow{6}{*}{CIFAR-10}                                               
 & F. Codebook~\cite{chen2019towards}, Nat. (Pretrain.)        & 90.40\%           & 0.00\%                                                                 & 0.00\%                                                                   \\
 
 & F. Codebook~\cite{chen2019towards}, Nat. (Retrain.)        & 93.06\%           & 0.02\%                                                                 & 0.09\%                                                                   \\
 
   & F. Codebook~\cite{chen2019towards}, ATTA (Retrain.)           & 85.76\%           & 27.53\%                                                                & 53.18\%                                                                  \\ 
                                                     
                                                                          & Ours: AC + AB, Nat. (Pretrain.)          & 78.54\%           & 0.04\%                                                                 & 0.40\%                                                                   \\
                                                                         
                                                                          & Ours: AC + AB, Nat. (Retrain.)          & 91.11\%           & 0.75\%                                                                 & 2.46\%                                                                   \\
                                                                          
                                                                                                                                                                      & Ours: AC + AB, ATTA (Pretrain.)             & 82.11\%           & 24.92\%                                                                & 47.73\%                                                                  \\\midrule
\multirow{6}{*}{GTSRB}                                                   & F. Codebook~\cite{chen2019towards}, Nat. (Pretrain.)        & 95.71\%           & 17.19\%                                                                & 25.91\%                                                                  \\
 & F. Codebook~\cite{chen2019towards}, Nat. (Retrain.)        & 95.53\%           & 19.34\%                                                                & 27.66\%                                                                  \\
 & F. Codebook~\cite{chen2019towards}, ATTA (Retrain.)           & 88.08\%           & 57.83\%                                                                & 76.02\%                                                                  \\ 

                                                                          & Ours: AC + AB, Nat. (Pretrain.)          & 92.27\%           & 18.21\%                                                                & 31.21\%                                                                  \\
                                                                         
                                                                          & Ours: AC + AB, Nat. (Retrain.)          & 94.32\%           & 15.77\%                                                                & 32.40\%                                                                  \\
                                                                         
                                                                            & Ours: AC + AB, ATTA (Pretrain.)             & 89.22\%           & 57.20\%                                                                & 71.93\%                                                                  \\
                                                    \midrule
\multirow{6}{*}{RESISC45}                                                    & F. Codebook~\cite{chen2019towards}, Nat. (Pretrain.)        & 94.62\%           & 0.00\%                                                                 & 0.00\%                                                                   \\
                    & F. Codebook~\cite{chen2019towards}, Nat. (Retrain.)        & 95.51\%           & 0.00\%                                                                 & 0.00\%                                                                   \\
                    
                     & F. Codebook~\cite{chen2019towards}, ATTA (Retrain.)           & 74.13\%           & 8.22\%                                                                 & 43.71\%                                                                  \\ 
                                                   
                    & Ours: AC + AB, Nat. (Pretrain.)          & 54.11\%           & 2.88\%                                                                 & 1.60\%                                                                   \\
                                                                       
                                                                          & Ours: AC + AB, Nat. (Retrain.)          & 92.11\%           & 6.64\%                                                                 & 3.42\%                                                                   \\
                                                                          
                                                                                                 & Ours: AC + AB, ATTA (Pretrain.)             & 62.02\%           & 20.04\%                                                                & 27.22\%                                                                  \\
                                                 \midrule
\multirow{6}{*}{Imagenette}  
 & F. Codebook~\cite{chen2019towards}, Nat. (Pretrain.)        & 98.29\%           & 0.79\%                                                                 & 0.66\%                                                                   \\
 & F. Codebook~\cite{chen2019towards}, Nat. (Retrain.)        & 98.83\%           & 1.48\%                                                                 & 1.07\%                                                                   \\
 
  & F. Codebook~\cite{chen2019towards}, ATTA (Retrain.)           & 89.55\%           & 65.68\%                                                                & 75.29\%                                                                  \\
 & Ours: AC + AB, Nat. (Pretrain.)          & 95.97\%           & 17.45\%                                                                & 20.41\%                                                                  \\
                                                                         
                                                                          & Ours: AC + AB, Nat. (Retrain.)          & 98.55\%           & 24.66\%                                                                & 30.85\%                                                                  \\

                                                                           & Ours: AC + AB, ATTA (Pretrain.)             & 90.93\%           & 65.89\%                                                                & 74.93\%                                                                  \\\bottomrule
\end{tabular}
\end{table}
\begin{table}
\centering
\caption{Square Attack~\cite{andriushchenko2020square} results on 500 samples for the results provided in Table~\ref{tab:wb_table_complex}. Square Attack does not significantly reduce the robustness on any of these models.}\label{tab:square1}
\begin{tabular}{clcc}
\toprule
\textbf{Dataset}                                                          & \multicolumn{1}{c}{\textbf{Model}} & \textbf{\begin{tabular}[c]{@{}c@{}}Robustness\\ ($L_2$ Square\\ on 500 Samples)\end{tabular}} & \textbf{\begin{tabular}[c]{@{}c@{}}Robustness\\ ($L_\infty$ Square\\ on 500 Samples)\end{tabular}} \\ \midrule

\multirow{4}{*}{CIFAR-10}                                                 & Baseline                               & 62.2\%                                                                                     & 62.6\%                                                                                       \\
                                            
                                                                          & Fixed Codebook~\cite{chen2019towards} (Pretrained)           & 62.0\%                                                                                     & 58.8\%                                                                                       \\
                                           & Ours: AC (Retrained) & 64.0\%                                                                                    & 65.0\%                                                                                        \\                                & Ours: AC + AB (Retrained)             & 63.6\%                                                                                     & 56.2\%                                                                                       \\
                                                                                                        
                                                                          \midrule

\multirow{4}{*}{RESISC45}                                                 & Baseline                               & 46.8\%                                                                                     & 63.8\%                                                                                       \\
                                                                         
                                                                          & Fixed Codebook~\cite{chen2019towards} (Pretrained)           & 53.2\%                                                                                     & 63.8\%                                                                                       \\
                                            & Ours: AC (Retrained) & 63.0\%                                                                                    &  60.4\%                                                                                        \\                                & Ours: AC + AB (Retrained)             & 60.8\%                                                                                     & 50.6\%                                                                                       \\
                                                                        
                                                                          \midrule
\multirow{4}{*}{Imagenette}                                               & Baseline                               & 88.6\%                                                                                     & 89.2\%                                                                                       \\
                                                                         
                                                                          & Fixed Codebook~\cite{chen2019towards} (Pretrained)           & 88.6\%                                                                                     & 88.4\%                                                                                       \\
                                                             & Ours: AC (Retrained) & 91.4\%                                                                                    &  91.0\%                                                                                        \\                                            & Ours: AC + AB (Retrained)             & 91.6\%                                                                                     & 91.4\%                                                                                       \\
                                                                         
                                                                          \bottomrule
\end{tabular}
\end{table}
\begin{table}
\centering
\caption{Square Attack~\cite{andriushchenko2020square} results on 500 samples for the results provided in Table~\ref{tab:wb_table_simple}. Square Attack does not significantly reduce the robustness on any models with \alg involved, but does significantly reduce the $L_2$ robustness of the ATTA + Chen~\etal models on MNIST and Fashion-MNIST.}\label{tab:square2}
\begin{tabular}{clcc}
\toprule
\textbf{Dataset}                                                          & \multicolumn{1}{c}{\textbf{Model}} & \textbf{\begin{tabular}[c]{@{}c@{}}Robustness\\ ($L_2$ Square\\ on 500 Samples)\end{tabular}} & \textbf{\begin{tabular}[c]{@{}c@{}}Robustness\\ ($L_\infty$ Square\\ on 500 Samples)\end{tabular}} \\ \midrule
\multirow{6}{*}{MNIST}                                                    & Baseline                               & 0.8\%                                                                                      & 83.8\%                                                                                       \\

                                                                          & Fixed Codebook~\cite{chen2019towards} (Pretrained)           & 13.0\%                                                                                     & 87.0\%                                                                                       \\
                                                & Ours: AC (Pretrained) &  11.0\%                                                                                     &  86.8\%                                                                                        \\                           & Ours: AC (Retrained) &  22.0\%                                                                                     &  90.2\%                                                                                        \\
                                        & Ours: AC + AB (Pretrained)            & 11.6\%                                                                                     & 86.0\%                                                                                       \\                                   & Ours: AC + AB (Retrained)             & 16.6\%                                                                                     & 89.8\%                                                                                       \\ 
                                                                           \midrule
\multirow{6}{*}{\begin{tabular}[c]{@{}c@{}}Fashion-\\ MNIST\end{tabular}} & Baseline                               & 29.8\%                                                                                     & 71.6\%                                                                                       \\
                                                                         
                                                                          & Fixed Codebook~\cite{chen2019towards} (Pretrained)           & 27.8\%                                                                                     & 66.6\%                                                                                       \\
                           & Ours: AC (Pretrained) & 31.8\%                                                                                    &  68.4\%                                                                                        \\                         & Ours: AC (Retrained) & 47.0\%                                                                                    &  69.4\%                                                                                        \\  
                           & Ours: AC + AB (Pretrained)             & 34.2\%                                                                                     & 67.8\%                                                                                       \\& Ours: AC + AB (Retrained)             & 50.2\%                                                                                     & 68.0\%                                                                                       \\ 
                                                                           \midrule

\multirow{6}{*}{GTSRB}                                                    & Baseline                               & 59.0\%                                                                                     & 81.0\%                                                                                       \\
                                                                         
                                                                          & Fixed Codebook~\cite{chen2019towards} (Pretrained)          & 66.2\%                                                                                     & 82.6\%                                                                                       \\
                    & Ours: AC (Pretrained) & 65.4\%                                                                                     & 83.4\%                                                                                        \\                            & Ours: AC (Retrained) & 65.0\%                                                                                     & 79.2\%                                                                                        \\    
                    & Ours: AC + AB (Pretrained)             & 60.0\%                                                                                     & 81.0\%                                                                                       \\
                    & Ours: AC + AB (Retrained)             & 62.0\%                                                                                     & 73.8\%                                                                                       \\ 
                                                                         
                        \bottomrule
\end{tabular}
\end{table}
\begin{table}
\centering
\caption{Square Attack~\cite{andriushchenko2020square} results on 500 samples on settings presented in Table~\ref{tab:supp_c}. Square Attack does not significantly reduce the robustness of models with \alg, although some models with Chen~\etal, such as the Natural Retrained with Chen, experience significant reduction. Our \alg adaptive attacks perform particularly well on some settings such as the Natural + EF at Test Time settings for RESISC45 and Imagenette.}\label{tab:square3}
\begin{tabular}{@{}clcc@{}}
\toprule
\textbf{Dataset}                                                          & \multicolumn{1}{c}{\textbf{Model}} & \textbf{\begin{tabular}[c]{@{}c@{}}Robustness\\ ($L_2$ Square \\ on 500 Samples)\end{tabular}} & \textbf{\begin{tabular}[c]{@{}c@{}}Robustness\\ ($L_\infty$ Square \\ on 500 Samples)\end{tabular}} \\ \midrule
\multirow{6}{*}{MNIST}                                                     & F. Codebook~\cite{chen2019towards}, Nat. (Pretrain.)        & 16.2\%                                                                    & 78.0\%                                                                      \\
                                             & F. Codebook~\cite{chen2019towards}, Nat. (Retrain.)        & 2.4\%                                                                     & 1.6\%                                                                       \\
                                            & F. Codebook~\cite{chen2019towards}, ATTA (Retrain.)           & 23.2\%                                                                    & 90.2\%                                                                      \\   
                                             & Ours: AC + AB, Nat. (Pretrain.)          & 13.8\%                                                                    & 73.4\%                                                                      \\
                                                                         
                                                                          & Ours: AC + AB, Nat. (Retrain)         & 18.6\%                                                                    & 75.2\%                                                                      \\

                                                                                                                                                    & Ours: AC + AB, ATTA (Pretrain.)            & 11.6\%                                                                                     & 86.0\%                                                                                       \\
                                                                                                                                                    \midrule
\multirow{6}{*}{\begin{tabular}[c]{@{}c@{}}Fashion-\\ MNIST\end{tabular}}      
 & F. Codebook~\cite{chen2019towards}, Nat. (Pretrain.)        & 9.6\%                                                                     & 42.0\%                                                                      \\
 
  & F. Codebook~\cite{chen2019towards}, Nat. (Retrain.)        & 4.8\%                                                                     & 7.2\%                                                                       \\
  
   & F. Codebook~\cite{chen2019towards}, ATTA (Retrain.)           & 20.8\%                                                                    & 69.8\%                                                                      \\ 

& Ours: AC + AB, Nat. (Pretrain.)          & 14.6\%                                                                    & 21.2\%                                                                      \\
                                                                         
                                                                          & Ours: AC + AB, Nat. (Retrain)          & 15.8\%                                                                    & 24.2\%                                                                      \\

                                                                           & Ours: AC + AB, ATTA (Pretrain.)             & 34.2\%                                                                                     & 67.8\%                                                                                       \\
                                                                           \midrule
\multirow{6}{*}{CIFAR-10}                                                 & F. Codebook~\cite{chen2019towards}, Nat. (Pretrain.)        & 32.2\%                                                                    & 0.0\%                                                                       \\
                              & F. Codebook~\cite{chen2019towards}, Nat. (Retrain.)        & 45.6\%                                                                    & 3.6\%                                                                       \\       & F. Codebook~\cite{chen2019towards}, ATTA (Retrain.)           & 45.0\%                                                                    & 50.0\%                                                                      \\          
                                                                          & Ours: AC + AB, Nat. (Pretrain.)          & 19.2\%                                                                    & 2.6\%                                                                       \\
                                                                                                       & Ours: AC + AB, Nat. (Retrain)          & 43.0\%                                                                    & 22.4\%                                                                      \\

                                                                                                        & Ours: AC + AB, ATTA (Pretrain.)             & 59.6\%                                                                                     & 58.0\%                                                                                       \\\midrule
\multirow{6}{*}{GTSRB}                                                   & F. Codebook~\cite{chen2019towards}, Nat. (Pretrain.)        & 50.4\%                                                                    & 42.\%                                                                       \\
 & F. Codebook~\cite{chen2019towards}, Nat. (Retrain.)        & 50.8\%                                                                    & 42.2\%                                                                      \\
 
 & F. Codebook~\cite{chen2019towards}, ATTA (Retrain.)           & 67.4\%                                                                    & 78.0\%                                                                      \\ 
                                                                          & Ours: AC + AB, Nat. (Pretrain.)          & 44.2\%                                                                    & 46.4\%                                                                      \\
                                                                         
                                                                          & Ours: AC + AB, Nat. (Retrain)         & 44.0\%                                                                    & 55.6\%                                                                      \\

                                                                           & Ours: AC + AB, ATTA (Pretrain.)             & 60.0\%                                                                                     & 81.0\%                                                                                       \\
                                                                           \midrule
\multirow{6}{*}{RESISC45}                                                                               & F. Codebook~\cite{chen2019towards}, Nat. (Pretrain.)        & 12.4\%                                                                    & 4.2\%                                                                       \\
& F. Codebook~\cite{chen2019towards}, Nat. (Retrain.)        & 9.8\%                                                                     & 4.2\%                                                                       \\

     & F. Codebook~\cite{chen2019towards}, ATTA (Retrain.)           & 33.6\%                                                                    & 39.6\%                                                                      \\                                                                      & Ours: AC + AB, Nat. (Pretrain.)          & 9.2\%                                                                     & 3.0\%                                                                       \\
                                           
                                                                          & Ours: AC + AB, Nat. (Retrain)         & 28.8\%                                                                    & 12.0\%                                                                      \\

                                                                           & Ours: AC + AB, ATTA (Pretrain.)             & 37.8\%                                                                                     & 33.8\%                                                                                       \\\midrule
\multirow{6}{*}{Imagenette}                                              & F. Codebook~\cite{chen2019towards}, Nat. (Pretrain.)        & 76.4\%                                                                    & 57.4\%                                                                      \\

 & F. Codebook~\cite{chen2019towards}, Nat. (Retrain.)        & 84.8\%                                                                    & 66.0\%                                                                      \\
                    & F. Codebook~\cite{chen2019towards}, ATTA (Retrain.)           & 80.6\%                                                                    & 83.0\%                                                                      \\                                                        & Ours: AC + AB, Nat. (Pretrain.)          & 74.4\%                                                                    & 61.2\%                                                                      \\
                                                                          
                                                                          & Ours: AC + AB, Nat. (Retrain)          & 88.2\%                                                                    & 81.0\%                                                                      \\

                                                                           & Ours: AC + AB, ATTA (Pretrain.)             & 85.6\%                                                                                     & 85.0\%                                                                                       \\\bottomrule
\end{tabular}
\end{table}

\section{Ablations}\label{supp:ablate}
We include results on just adaptive blurring and just adaptive color reduction in Tables~\ref{tab:supp_e} and~\ref{tab:supp_e_2}. We only test with the setting that adversarially trains the model with the transform.

We find that the full \alg appears better than just adaptive blurring in every case except for GTSRB, which had not improved over the baseline anyways. For adaptive color reduction, we can see that on the higher resolution datasets the full transform does the best on natural (88.24\% vs 83.44\% on RESISC45, 95.34\% vs. 94.75\% on Imagenette), $L_2$ PGD robustness (52.44\% vs. 30.80\% on RESISC45, 72.28\% vs. 71.62\% on Imagenette), and $L_\infty$ PGD robustness (53.33\% vs. 45.24\% on RESISC45, 80.71\% vs. 79.90\% on Imagenette). The performance increase is particularly clear on RESISC45. The Imagenette gaps are not large, but that may be due to its smaller $\epsilon$ perturbation values. CIFAR-10 also benefits from the full pipeline on $L_2$ attacks compared to the adversarially trained model with \alg (29.45\% vs. 38.46\%). 

It is perhaps not too surprising that on datasets with low resolution and well split colors already (i.e., MNIST, Fashion-MNIST, GTSRB) just adaptive color reduction performs similarly to the full transform. We also find that Square Attack~\cite{andriushchenko2020square} does not drastically reduce the accuracy compared to PGD, suggesting that the adaptive attacks are reasonable.

\begin{table}
\centering
\caption{Results comparing the full transform with just adaptive blurring and just adaptive color reduction. The full transformation shows clear benefits over just adaptive blurring and helps significantly on CIFAR-10 and RESISC45.}\label{tab:supp_e}
\begin{tabular}{clccc}
\toprule
\textbf{Dataset}                                                          & \multicolumn{1}{c}{\textbf{Model}}                                       & \textbf{Accuracy} & \textbf{\begin{tabular}[c]{@{}c@{}}Robustness\\ ($L_2$ PGD)\end{tabular}} & \textbf{\begin{tabular}[c]{@{}c@{}}Robustness\\ ($L_\infty$ PGD)\end{tabular}} \\ \midrule
\multirow{3}{*}{MNIST}                                                    & Ours: AC + AB                                                                  & 98.43\%           & 0.00\%                                                                                     & 91.00\%                                                                                         \\
                                                                          & Ours: AB                                                       & 96.70\%           & \textbf{0.02\%}                                                                                     & 84.97\%                                                                                       \\
                                                                          & Ours: AC & 98.51\%           & 0.00\%                                                                                      & \textbf{91.73\%}                                                                                        \\ \midrule
\multirow{3}{*}{\begin{tabular}[c]{@{}c@{}}Fashion-\\ MNIST\end{tabular}} & Ours: AC + AB                                                                  & 82.06\%           & \textbf{23.37\%}                                                                                   & 71.54\%                                                                                       \\
                                                                          & Ours: AB                                                       & 81.86\%           & 20.24\%                                                                                    & 69.15\%                                                                                        \\
                                                                          & Ours: AC & 82.22\%           & 22.06\%                                                                                    & \textbf{71.88\%}                                                                                       \\ \midrule
\multirow{3}{*}{CIFAR-10}                                                 & Ours: AC + AB                                                                  & 84.52\%           & \textbf{38.46\%}                                                                                     & \textbf{51.97\%}                                                                                        \\
                                                                          & Ours: AB                                                       & 83.78\%           & 23.15\%                                                                                    & 50.05\%                                                                                        \\
                                                                          & Ours: AC & 85.27\%           & 29.45\%                                                                                    & 50.14\%                                                                                       \\ \midrule
\multirow{3}{*}{GTSRB}                                                    & Ours: AC + AB                                                                  & 90.32\%           & 57.71\%                                                                                    & 73.25\%                                                                                      \\
                                                                          & Ours: AB                                                       & 92.15\%           & 57.17\%                                                                                   & 76.32\%                                                                                         \\
                                                                          & Ours: AC & 91.86\%           & \textbf{58.7\%}                                                                                    & \textbf{76.45\% }                                                                                       \\ \midrule
\multirow{3}{*}{RESISC45}                                                 & Ours: AC + AB                                                                  & 88.24\%           & \textbf{52.44\%}                                                                                    & \textbf{53.33\%}                                                                                    \\
                                                                          & Ours: AB                                                       & 84.07\%           & 21.47\%                                                                                  & 41.38\%                                                                                       \\
                                                                          & Ours: AC & 83.44\%           & 30.80\%                                                                                  & 45.24\%                                                                                         \\ \midrule
\multirow{3}{*}{Imagenette}                                               & Ours: AC + AB                                                                 & 95.34\%           & \textbf{72.28\% }                                                                                  & \textbf{80.71\%}                                                                                    \\
                                                                          & Ours: AB                                                       & 94.45\%           & 66.50\%                                                                                    & 76.2\%                                                                                        \\
                                                                          & Ours: AC & 94.75\%           & 71.62\%                                                                                  & 79.90\%                                                                                        \\ \bottomrule
\end{tabular}
\end{table}

\begin{table}
\centering
\caption{Results comparing the full transform with just adaptive blurring and just adaptive color reduction on Square Attack.}\label{tab:supp_e_2}
\begin{tabular}{clccc}
\toprule
\textbf{Dataset}                                                          & \multicolumn{1}{c}{\textbf{Model}}                                       & \textbf{Accuracy} & \textbf{\begin{tabular}[c]{@{}c@{}}Robustness\\ ($L_2$ Square \\ on 500 \\ Samples)\end{tabular}} & \textbf{\begin{tabular}[c]{@{}c@{}}Robustness\\ ($L_\infty$ Square \\ on 500 \\ Samples)\end{tabular}} \\ \midrule
\multirow{3}{*}{MNIST}                                                    & Ours: AC + AB                                                                  & 98.43\%           & 16.6\%                                                                                     & 89.8\%                                                                                        \\
                                                                          & Ours: AB                                                       & 96.70\%           &  1.0\%                                                                                     &  83.0\%                                                                                        \\
                                                                          & Ours: AC & 98.51\%           &  22.0\%                                                                                     &  90.2\%                                                                                        \\ \midrule
\multirow{3}{*}{\begin{tabular}[c]{@{}c@{}}Fashion-\\ MNIST\end{tabular}} & Ours: AC + AB                                                                  & 82.06\%           &  50.2\%                                                                                    &  68.0\%                                                                                        \\
                                                                          & Ours: AB                                                       & 81.86\%           & 36.2\%                                                                                    &  69.2\%                                                                                        \\
                                                                          & Ours: AC & 82.22\%           &   47.0\%                                                                                    &  69.4\%                                                                                        \\ \midrule
\multirow{3}{*}{CIFAR-10}                                                 & Ours: AC + AB                                                                  & 84.52\%           & 63.6\%                                                                                    & 56.2\%                                                                                        \\
                                                                          & Ours: AB                                                       & 83.78\%           & 63.2\%                                                                                    & 59.0\%                                                                                        \\
                                                                          & Ours: AC & 85.27\%           &  64.0\%                                                                                    & 65.0\%                                                                                        \\ \midrule
\multirow{3}{*}{GTSRB}                                                    & Ours: AC + AB                                                                  & 90.32\%           & 62.0\%                                                                                    &  73.8\%                                                                                        \\
                                                                          & Ours: AB                                                       & 92.15\%           &  59.2\%                                                                                    &  77.4\%                                                                                        \\
                                                                          & Ours: AC & 91.86\%           &  65.0\%                                                                                     & 79.2\%                                                                                        \\ \midrule
\multirow{3}{*}{RESISC45}                                                 & Ours: AC + AB                                                                  & 88.24\%           &  60.8\%                                                                                    &  50.6\%                                                                                        \\
                                                                          & Ours: AB                                                      & 84.07\%           &  52.0\%                                                                                    &  64.2\%                                                                                        \\
                                                                          & Ours: AC & 83.44\%           &  63.0\%                                                                                    &  60.4\%                                                                                        \\ \midrule
\multirow{3}{*}{Imagenette}                                               & Ours: AC + AB                                                                  & 95.34\%           &  91.6\%                                                                                    &  91.4\%                                                                                        \\
                                                                          & Ours: AB                                                       & 94.45\%           &  88.4\%                                                                                    & 90.4\%                                                                                         \\
                                                                          & Ours: AC & 94.75\%           &  91.4\%                                                                                    &  91.0\%                                                                                        \\ \bottomrule
\end{tabular}
\end{table}

\section{Analysis}
\subsection{GTSRB Results}\label{sec:supp_gtsrb}

We now analyze our GTSRB results as one of the datasets where \alg did not improve performance in any setting. We observe that the baselines already perform fairly well, making further improvements difficult. In addition, we found several examples of attacks that would cause enough confusion to possibly mess up a human's decision because of their blurry nature and inherent reliance on particular features at a couple of key spots.

\begin{figure}[t]
    \centering
    {\makebox[0.75in]{\includegraphics[width=0.75in]{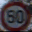}}}%
    \hfil
    {\makebox[0.75in]{\includegraphics[width=0.75in]{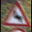}}}%
    \hfil
    {\makebox[0.75in]{\includegraphics[width=0.75in]{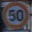}}}%
    \hfil
    {\makebox[0.75in]{\includegraphics[width=0.75in]{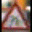}}}%
    \caption{Example attacks on \alg GTSRB model where the attacked image appears to be a different class or ambigious. \textbf{Left}: $L_\infty$ attack that looks like it may be a 50 km/hr. Actual GT label: 80 km/hr. \textbf{Left Center}: $L_\infty$ attack that has little detail, and could feasibly be a wild animal crossing sign. Actual GT label: Winding Road. \textbf{Center Right}: $L_2$ attack that looks like a 50 km/hr. Actual GT label: 30 km/hr. \textbf{Right}: $L_2$ attack that has very little detail inside. Actual GT label: Bike crossing. }
    \label{fig:gtsrb}
\end{figure}

We show a couple of $L_2$ and $L_\infty$ examples in Fig.~\ref{fig:gtsrb} where the attack wiped away some of the distinguishing features in key parts of image (such as the numbering on speed limit signs). Such attacks will be hard for any defense to defend against, and thus explains why \alg appears to perform better on datasets with more naturally occurring images such as CIFAR-10 or RESISC45 that do not have as much dependency on small, specific features.

\subsection{Perceptibility}\label{sec:supp_percept}
We now test if attacks on \alg models are more perceptible than vanilla ATTA models, given that our transformation makes it harder to attack without changing regions of color-wise and spatial locality together. We first measure the SSIM~\cite{wang2004image} of $L_\infty$ attacked images from the pretrained ATTA baseline and our retrained ATTA model with \alg against the original, where higher SSIM means more similarity to the original.

\begin{figure}[t]
    \centering
    {\makebox[1.4in]{\includegraphics[width=1.1in]{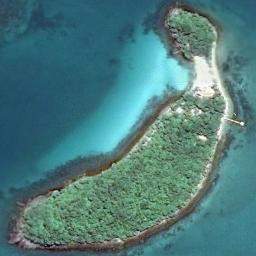}}}%
    \hfil
    {\makebox[1.4in]{\includegraphics[width=1.1in]{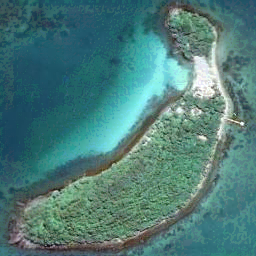}}}%
    \hfil
    {\makebox[1.4in]{\includegraphics[width=1.1in]{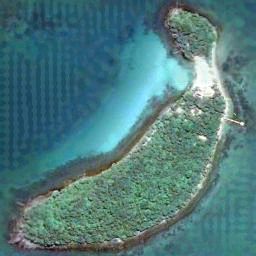}}}%
    \caption{SSIM does not necessarily capture the change in perceptibility on \alg model attacks. \textbf{Left}: original image. \textbf{Center}: vanilla ATTA model attack. SSIM: 0.893. \textbf{Right}: \alg trained ATTA model attack. SSIM: 0.868.}
    \label{fig:ssim}
\end{figure}

We tested on the 4 color datasets and found on average there was not a significant change from the average SSIM on pretrained ATTA attacks versus the average SSIM on retrained ATTA with \alg attacks (0.946 vs. 0.944 for CIFAR-10, 0.927 vs. 0.933 for GTSRB, 0.899 vs. 0.877 for RESISC45, 0.955 vs. 0.963 for Imagenette). The low resolution is a likely explanation for CIFAR-10 and GTSRB, and for the higher resolution datasets, we qualitatively find that the \alg attacks often exhibit grid like blob perturbations that are not necessarily reflected in SSIM in large areas of similar color. We show in Fig.~\ref{fig:ssim} an example of a retrained ATTA with \alg attack with clear perturbations but a similar SSIM score as a pretrained ATTA baseline attack. This shows that neither SSIM nor $L_\infty$ necessarily capture the true increase of perceptibility. We leave more exploration to future work.

\end{document}